\documentclass[journal]{IEEEtran}

\ifCLASSINFOpdf
\else
   \usepackage[dvips]{graphicx}
\fi
\usepackage{url}

\usepackage{graphicx}
\usepackage{subcaption}
\usepackage{amsmath}
\usepackage{cite}
\usepackage{floatrow}
\usepackage{gensymb}
\usepackage{amsfonts}
\usepackage{etoolbox}
\usepackage{enumerate}
\usepackage{multirow}
\usepackage{romannum}
\usepackage{booktabs}
\usepackage{graphicx}
\usepackage{balance}
\usepackage{siunitx}
\usepackage{algorithmic}
\usepackage[linesnumbered,ruled,vlined]{algorithm2e}
\usepackage{soul}
\usepackage{lipsum}% http://ctan.org/pkg/lipsum
\usepackage{graphicx}% http://ctan.org/pkg/graphicx
\usepackage{stfloats}
\usepackage [autostyle, english = american]{csquotes}
\usepackage{amsmath}
\DeclareMathOperator*{\minimize}{minimize}
\usepackage{amsmath}
\usepackage{ mathrsfs }
\DeclareMathOperator{\Tr}{Tr}
\DeclareMathOperator{\Relu}{Relu}
\MakeOuterQuote{"}
\usepackage{caption}

\begin{document}

\title{A Deep-Unfolded Spatiotemporal RPCA Network For L+S Decomposition}

\author{Shoaib Imran, Muhammad Tahir \IEEEmembership{Senior Member, IEEE}, Zubair Khalid \IEEEmembership{Senior Member, IEEE}, and Momin Uppal \IEEEmembership{Senior Member, IEEE}}
%\thanks{This paragraph of the first footnote will contain the date on which you submitted your paper for review. It will also contain support information, including sponsor and financial support acknowledgment. For example, ``This work was supported in part by the U.S. Department of Commerce under Grant BS123456.'' }
%\thanks{The next few paragraphs should contain the authors' current affiliations, including current address and e-mail. For example, F. A. Author is with the National Institute of Standards and Technology, Boulder, CO 80305 USA (e-mail: author@boulder.nist.gov).}
%\thanks{S. B. Author, Jr., was with Rice University, Houston, TX 77005 USA. He is now with the Department of Physics, Colorado State University, Fort Collins, CO 80523 USA (e-mail: author@lamar.colostate.edu).}
%}
%\markboth{Journal of \LaTeX\ Class Files, Vol. 14, No. 8, August 2015}
%{Shell \MakeLowercase{\textit{et al.}}: Bare Demo of IEEEtran.cls for IEEE Journals}
\maketitle

\begin{abstract}
	Low-rank and sparse decomposition  based methods find their use in many applications involving background modeling such as clutter suppression and object tracking. While Robust Principal Component Analysis (RPCA) has achieved great success in performing this task, it can take hundreds of iterations to converge and its performance decreases in the presence of different phenomena  such as occlusion, jitter and fast motion. The recently proposed deep unfolded networks, on the other hand, have demonstrated better accuracy and improved convergence over both their iterative equivalents as well as over other neural network architectures. In this work, we propose a novel deep unfolded spatiotemporal RPCA (DUST-RPCA) network, which explicitly takes advantage of the spatial and temporal continuity in the low-rank component. Our experimental results on the moving MNIST dataset indicate that DUST-RPCA gives better accuracy  when compared with the existing state of the art deep unfolded RPCA networks.
\end{abstract}
\IEEEpeerreviewmaketitle
\section{Introduction}
Robust Principal Component Analysis (RPCA) has become a popular choice in a range of applications including, but not limited to, computer vision \cite{8425659}, speech enhancement \cite{7740039} and data analysis \cite{Wright-Ma-2022}. In fact, different variants of RPCA exist for various problem settings \cite{8425660}. RPCA decomposes a data matrix $D$, the columns of which are constructed from vectorized data frames, into a sparse matrix $S$, which models the foreground and a low-rank matrix $L$, which models the background. RPCA is usually formulated on the Principal Component Pursuit (PCP) model. However, the PCP model's performance decreases in the presence of complex occurrences such as background clutter, occlusion and jitter. In such scenarios, a number of methods \cite{7036101}, \cite{DBLP:journals/tcyb/ZhouLBY18} and \cite{7396944} have been proposed which exploit the robust correlation between the data frames (columns of the input matrix $D$). As an example, the authors in \cite{8017547},  observe that these phenomena usually span multiple low-dimensional manifolds in the low-rank representation. Therefore, by incorporating spatial and temporal constraints in the PCP framework,  one can control the coherency on these low-dimensional manifolds, resulting in better background-foreground separation accuracy. However, being an optimization method, it can take many iterations to converge with each iteration involving computationally expensive singular value decomposition (SVD) operation on high-dimensional data.

Deep Neural Networks (DNNs), such as ResNet \cite{inproceedings_o}, on the other hand, have shown to be much more computationally efficient than optimization based subspace methods while also being robust to occlusion, clutter and fast motion. However, DNNs often require large training datasets and extreme parameterization \cite{shlezinger2022model}. This together with the abstractness of DNNs essentially make them black boxes and non-interpretable.

The recently proposed deep unfolded neural networks, designed by mapping the iterative steps of an optimization algorithm into DNN layers with learnable parameters, aim at solving the interpretability and large training dataset issues of DNNs while avoiding the high computational cost that is associated with traditional iterative optimization. Building upon this approach, the CORONA  network \cite{8836615} solved the low-rank and sparse $(L+S)$ decomposition problem  using convolution kernels in its unfolding process. As a result, it attained better performance as compared to  its iterative equivalent and to other prominent DNNs, such as ResNet. The CORONA network's model, however, does not explicitly take advantage of the high correlation between successive frames of both the background and the foreground. The refRPCA \cite{Luong2021ADR} was proposed to solve this problem by incorporating side-information schemes in the shape of (i) a projection matrix $P$ to potentially represent the high correlation between successive frames of the sparse component $S$, and (ii) a reweighting matrix $Q$ to reweight the thresholds of the proximal operator. Both $P$ and $Q$ matrices are trainable parameters that are learnt by the network, making refRPCA more adaptive such that it outperforms the CORONA network. However, once refRPCA is trained, the reweighting matrix $Q$ remains invariant to the data matrix $D$, which may vary significantly from one sample to another. Moreover, refRPCA model doesn't exploit the high correlation between the low-rank background frames.

In this paper, we propose a novel deep unfolded spatiotemporal RPCA (DUST-RPCA) network which explicitly takes advantage of the high correlation between successive frames of the background. This is in contrast to the correlation between foreground frames which may vary significantly for various applications. Consequently, we use the masks of the sparse component, given by the computationally efficient GoDec algorithm \cite{10.5555/3104482.3104487}, to reweight the thresholds of the proximal operator \cite{9906418}. To exploit the high background correlation, we add constraints that ensure continuity in both spatial and temporal dimensions of the background, which as per the model of RPCA is low-rank. The spatial constraint encodes the similarities between the rows of the input matrix $D$, whereas the temporal constraint encodes the similarities between the columns. Our experiments on the moving MNIST dataset show that DUST-RPCA achieves better accuracy  than both CORONA and refRPCA.

\section{Problem Formulation}
\subsection{The Proposed Objective Function}
Consider a sequence of data frames, each denoted by $d_i$ for $\text{time instances } i=1,2,\cdots, q$  and each of size $m \times n$,  having their low-rank and sparse  components denoted as $l_i$ and $s_i$, respectively. The data sequence $d_i$ is arranged into a matrix $D$ of size $p\times q$, where $p=mn$. Similarly,  for sequences $l_i$ and $s_i$, we obtain their corresponding matrices $L$ and $S$, both of sizes $p\times q$. The RPCA, in its  PCP formulation, separates the data matrix $D$ into its low-rank component $L$ and the sparse component $S$,  as \cite{10.1145/1970392.1970395}
\begin{equation} \label{t1}
\min_{ L, S} \quad  \|L\|_* + \lambda \|S\|_1     \text{ s.t. }  {D} = {L} + {S}, 
\end{equation}where $\|.\|_{1}$ is the $\ell_{1}$ norm, $\|.\|_{*}$ is the nuclear norm and $\lambda$ is a tuning regularizer. The PCP formulation, however, shows degraded performance amidst background clutter, occlusion, jitter etc. However, these challenges can be addressed by incorporating Laplacian based graph regularization  in the PCP framework \cite{8017547}. Consequently, these structural constraints are used to incorporate the spatial and temporal similitudes by reformulating \eqref{t1} as
 \begin{gather}\label{t2}
\min_{ L, S} \quad \|L\|_* + \lambda \|W\circ S\|_1 +  \gamma_1\Tr(L^TA_sL)  + \\ \gamma_2\Tr(LA_tL^T)  \text{ s.t. } {D} = {L} + {S}, \notag
 \end{gather}
where $\Tr()$ denotes the trace of a matrix, $\circ$ denotes the Hadamard product, $A_s$ and $A_t$ are the spatial and temporal Laplacian matrices respectively, and $\gamma_1$ and $\gamma_2$  are the regularization parameters which balance the spatial and temporal prior information, respectively. In addition, we also reweight the elements of $S$, as done in \cite{9906418}, with a prior matrix $W= \Phi(\rho, \hat{W} )$, where  $\Phi(.)$ is the sigmoid function with a  gain of $\rho$ and $\hat{W}$ is the background mask obtained by inverting the foreground mask from the GoDec algorithm. GoDec is significantly more efficient than ADMM \cite{book} as it does not use time-consuming SVD operation for estimating the low-rank component. On average, the GoDec takes about 7\% of the time taken by the ADMM algorithm in giving a similar mask. 

The Laplacian matrices are computed as $ A_g = T_g - W_g$, where $g \in \{s,t\}$, and  $T_g$ and $W_g$ denote the corresponding adjacency and degree matrices respectively. The adjacency matrix $W_g$ is computed as 
 \begin{equation}\label{t3}
 W_g(i,j) = \exp\left( \frac{-\|x_i -x_j\|^2}{\tau_g  \cdot\kappa_g^2}\right)
 \end{equation}In case of $W_s$,  $x_i$ and $x_j$ denote the $i^{\text{th}}$ and $j^{\text{th}}$ rows whereas in case of $W_t$,  $x_i$ and $x_j$ denote the $i^{\text{th}}$ and $j^{\text{th}}$ columns of the input matrix $D$ respectively. On the other hand, $T_g$  is a diagonal matrix, with its diagonal entries composed of the sum of each row of $W_g$. The parameter $\kappa_s$ is the average distance between the rows whereas $\kappa_t$ is the average distance between the columns of the input matrix $D$ respectively. The parameter $\tau_g$ is a constant which is further used to tune the smoothness between the values of $W_g$. To handle $\| .\|_*$ and $\|.\|_{1}$ in \eqref{t2} separately, we  introduce two auxiliary matrices $U_1$ and $U_2$ to reformulate \eqref{t2} as \begin{equation} \label{t4}
\begin{gathered}
\min_{ L, S} \quad  \|L\|_* + \lambda \|W\circ S\|_1 +  \gamma_1\Tr(U_1^TA_sU_2)+\\  \gamma_2\Tr(U_2A_tU_2^T) 
\text{  s.t.  } D =L+S , L=U_1, L=U_2 
\end{gathered}
\end{equation}
\subsection{Subproblems for Spatiotemporal RPCA}
The augmented Lagrangian function associated with \eqref{t4} can be written as
\begin{gather} 
\mathscr{L}\left(L, S, U_1, U_2, Y_1, Y_2, Y_3 \right) = \|L\|_* + \lambda \|W \circ S \|_1  + \notag \\
\gamma_1\Tr(U_1^TA_sU_1) +\gamma_2\Tr(U_2A_tU_2^T)  + \Tr(Y_1 ^T\cdot(D-L-S)) \notag\\ + \Tr(Y_2^T\cdot(L-U_1)) +  \Tr(Y_3^T\cdot(L-U_2)) + \notag \\
\frac{\mu}{2}\left(\|D-L-S\|_F^2  + \|L-U_1\|_F^2 +\|L-U_2\|_F^2\right) \label{tt5}
\end{gather}where $\|.\|_F$ denotes the Frobenius norm, $Y_1$, $Y_2$ and $Y_3$ are the Lagrange multipliers of size $p \times q$, and $\mu$ is a regularization parameter that controls the penalty for violation of linear constraints. We solve \eqref{tt5} using alternating minimization \cite{NIPS2011_18997733}, which involves separately minimizing the augmented Lagrangian function. In  $k^{\text{th}}$ iteration, \eqref{tt5} is minimized with respect to each optimization variable while the remaining variables are kept constant. The sub-problems for updating $L$, $S$, $U_1$ and $U_2$ can be written as
\begin{equation}\label{t6}
\min_{U_1} \quad  \frac{\gamma_1}{\mu} \Tr(U_1^TA_sU_1) + \frac{1}{2} \|L-U_1 + \frac{Y_2}{\mu} \|^2_F,
\end{equation}\begin{equation}\label{t8}
\min_{U_2} \quad \frac{\gamma_2}{\mu} \Tr(U_2A_tU_2^T) + \frac{1}{2} \|L-U_2 + \frac{Y_3}{\mu} \|^2_F,
 \end{equation}\begin{gather}\label{t88}
\min_{L} \quad \frac{1}{\mu}\|L\|_* + \frac{1}{2}\|D-L-S+ U_1 + U_2 \\+\frac{\left(Y_1 + Y_2 + Y_3 \right)}{\mu}        \|_F^2, \notag
\end{gather}\begin{gather}\label{t9}
\min_{ S} \quad  \frac{\lambda}{\mu}\|W \circ S\|_1 + \frac{1}{2}\| D - L -S +\frac{Y_1}{\mu}\|_F^2. 
\end{gather}The solutions to sub-problems (6) and (7), for updating $U_1$ and $U_2$ at iteration $k$, can be found in closed form as \begin{equation}\label{t10}
U_1^{k+1} = \left(2\gamma_1A_s +\mu^kI\right)^{-1} \left(\mu^kL^k +Y^k_2\right)
\end{equation}\begin{equation}\label{t11}
U_2^{k+1} =\left(\mu^kL^k +Y_3^k\right) \left(2\gamma_2A_t +\mu^kI\right)^{-1}
\end{equation}
\section{Unfolding Spatiotemporal RPCA}
Although there are different ways to unfold an iterative algorithm into a DNN such as described in \cite{8836615}, \cite{cai2021learned} and\cite{monga2021algorithm}, we follow the method of \cite{8836615} in unfolding \eqref{t88} and \eqref{t9}. Hence, we introduce measurement matrices $\{H_i\}_{i=1}^{4}$ into \eqref{t88} and \eqref{t9} as follows
 \begin{gather}\label{t12}
\minimize_{ L} \quad  \frac{1}{\mu}\|L\|_* + \frac{1}{2}\|D-H_1L-H_2S+ H_3U_1\\+ H_4U_2 +\frac{\left(Y_1 + Y_2 + Y_3 \right)}{\mu}        \|_F^2, \notag
\end{gather}\begin{gather}\label{t13}
\minimize_{S} \quad  \frac{\lambda}{\mu}\|W \circ S\|_1 + \frac{1}{2}\| D - H_1L -H_2S+  \frac{Y_1}{\mu}\|_F^2 . 
\end{gather}
While $\{H_i\}_{i=1}^{4}$ can ideally be set with respect to the application at hand, we take them to be identity matrices in our case since we have no prior knowledge of a more informed choice. Moreover, they are further abstracted to enable the architecture of DUST-RPCA by replacing them with learnable kernels in the deep unfolding steps. We solve \eqref{t12} and \eqref{t13} using ISTA\cite{doi:10.1137/080716542},  as
\begin{equation}\label{t14}
B^{k+1} = \text{prox}\left(B^k -\frac{1}{c}\nabla\left(B^k \right) \right),
\end{equation}where $B$ can be either $L$ or $S$, $c$ denotes the Lipschitz constant, prox(.) denotes the proximal operator \cite{6733349} and $\nabla$(.) denotes the gradient of the quadratic part with respect to $B$. The proximal operator for computing the low-rank component is the singular value thresholding operation  \cite{doi:10.1137/080738970}, $\Psi_{\alpha }({X})= {\mathbf{U}}\textrm {diag}(\Relu(\sigma _{i}-\alpha)) {\mathbf{V}}^{H}$, where $X$ has an SVD given by $X = U\Sigma V^T$, diag($y_i$) represents a diagonal matrix with its $i^\text{th}$ diagonal entry equal to $y_i$ and $\sigma_i$ represents the $i^{\text{th}}$ singular value of $X$. The proximal operator for computing the sparse component is the  soft-thresholding operation \cite{doi:10.1137/080716542},  $\Theta_\alpha(x)  = \text{sign}(x)\cdot \text{max}(|x|-\alpha,0)$. The thresholding value $\alpha$ at iteration $k$ for $\Theta_{\alpha}$ is kept as $\frac{\lambda^k}{\mu^k \cdot c}\cdot \Phi(\rho^k, \hat{w} )$, where $\hat{w}$ is the element of the matrix $\hat{W}$. This design leads to a distinct proximal operator for every input at each iteration. Using \eqref{t14}, the iterative steps of $L$ and $S$, after some algebraic manipulation, turn out to be
\begin{figure}
		\hspace{-0cm}
		\centering
		\includegraphics[width=0.8\linewidth,height=4.45cm]{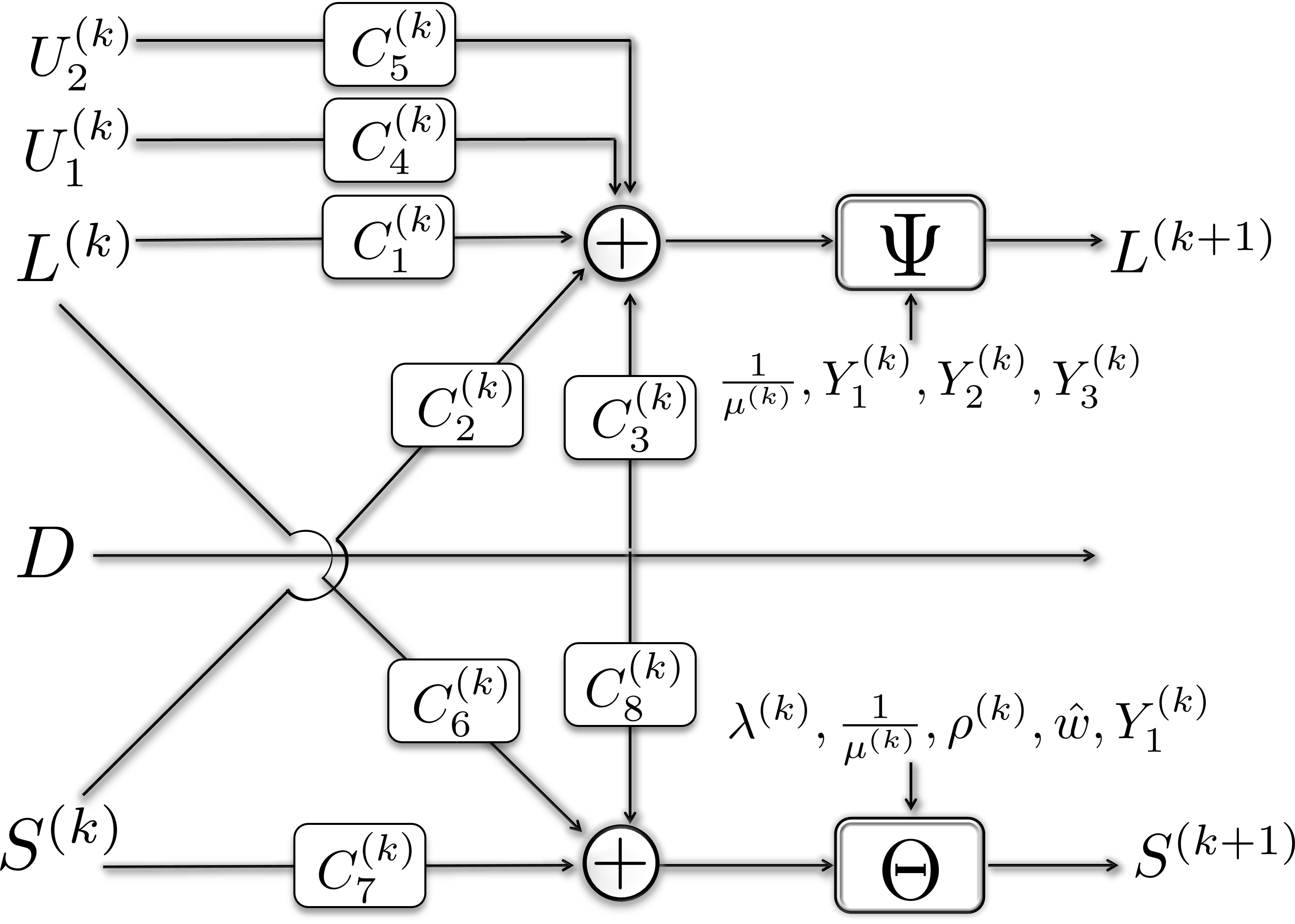}
		\caption{DUST-RPCA architecture} 	
		\label{fig:1}
	\end{figure}\begin{gather}\label{t15}
L^{k+1} = \Psi_{\frac{1}{\mu^k\cdot c}}
\Biggl\{
\left(I-\frac{1}{c}H_1^TH_1 \right)L^k +
\left(-H_1^TH_2 \right)S^k \\+ 
\left(H_1^T \right)D + \left(H_1^TH_3 \right)U_1^k + \left(H_1^TH_4\right)U_2^k + \notag\\ \left(H_1^T \right) \frac{Y_1^k + Y_2^k + Y_3^k}{\mu^k} \notag
\Biggl\},
\end{gather}\begin{gather}\label{t16}
S^{k+1} = \Theta_{\frac{\lambda^k}{\mu^k \cdot c}\cdot\Phi(\rho^k, \hat{w})}
\Biggl\{
\left(I-\frac{1}{c}H_2^TH_2 \right)S^k +\\
\left(-H_2^TH_1 \right)L^k + 
\left(H_2^T \right)D + \left(H_2^T \right) \frac{Y_1^k }{\mu^k}
\Biggl\}. \notag
\end{gather}
 We unroll \eqref{t15} and \eqref{t16} into the multi-layer neural network of DUST-RPCA by replacing the functions of the measurement matrices  with convolution kernels as follows
\begin{gather}\label{t17}
L^{k+1} = \Psi_{\frac{1}{\mu^k}}\Biggl\{C_1^k*L^k + C_2^k*S^k +C_3^k*D + \\C_4^k*U_1^k +  C_5^k*U_2^k + \frac{Y_1^k +Y_2^k + Y_3^k}{\mu^k} \notag
\Biggl\} 
\end{gather}\begin{align}\label{t18}
    {{S}}^{k+1}=&\Theta_{\frac{\lambda^k}{\mu^k}\cdot\Phi(\rho^k, \hat{w})} \left \{{ {{C}}_{6}^{k}* {{L}}^{k} + {{C}}_{7}^{k}* {{S}}^{k} + {{C}}_{8}^{k}* {{D}} +   {{\frac{Y_1^k}{\mu^k}}} }\right \}, 
\end{align} \begin{figure}
  \centering
  \captionsetup{justification=centering}
    \begin{subfigure}{0.17\textwidth}
    \captionsetup{justification=centering}
    \includegraphics[width=1.0\linewidth]{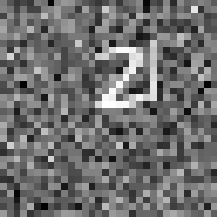}
    \caption{}
   \label{fig:3a}
  \end{subfigure}
  \begin{subfigure}{0.17\textwidth}
    \includegraphics[width=1.0\linewidth]{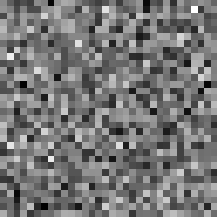}
    \caption{}
   \label{fig:3b}
  \end{subfigure}
    \begin{subfigure}{0.17\textwidth}
    \includegraphics[width=1.0\linewidth]{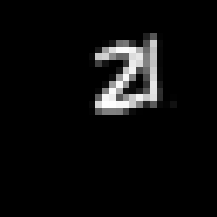}
    \caption{}
   \label{fig:3c}
  \end{subfigure}
    \begin{subfigure}{0.17\textwidth}
    \includegraphics[width=1.0\linewidth]{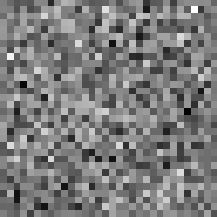}
    \caption{}
   \label{fig:3d}
  \end{subfigure}
    \begin{subfigure}{0.17\textwidth}
    \includegraphics[width=1.0\linewidth]{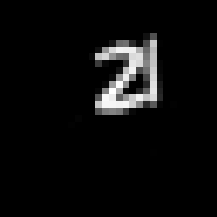}
    \caption{}
   \label{fig:3e}
  \end{subfigure}
  \caption{Separation results of DUST-RPCA: (a) Input image (b) Background groundtruth (c) Foreground groundtruth (d) Background recovered  (e) Foreground recovered }
  \label{fig:3}
\end{figure}where  $*$ denotes the convolution operaton, $C_1^k, C_2^k...C_8^k$ are the convolution kernels learned in the $k^{\text{th}}$ layer of DUST-RPCA network. The Lagrange multipliers of  $Y_1^k, Y_2^k$ and $Y_3^k$ in \eqref{t10}, \eqref{t11}, \eqref{t17} and \eqref{t18} are also learned by the network at every layer together with the parameters $ 
1/\mu^k, \lambda^k$ and $\rho^k$. Fig. \ref{fig:1} shows the $k^{\text{th}}$ layer of the architecture of DUST-RPCA.
\section{Experimental Evaluations}
\subsection{Experiment Settings}
We compare the performance of DUST-RPCA with CORONA and refRPCA in the foreground-background separation task on the moving MNIST dataset \cite{10.5555/3045118.3045209}, which contains 10,000 moving digits sequences, each 20 frames long. Each of these frames is resized to $32\times 32$ pixels and then stacked as columns to create the sparse matrix $S$ of size $1024 \times 20$. Furthermore, we sample from a standard Gaussian distribution to generate matrices $U\in \mathbb{R}^{1024 \times r}$ and $V\in \mathbb{R}^{20 \times r}$, with the rank $r$ set to 5. These matrices are in turn used to generate the low-rank matrix $L=UV^T$. We superimpose the MNIST digits in $S$ on the low-rank matrix $L$ to create the data matrix $D$. Finally, with the data sequences normalized to the range $[0,1]$, we split the dataset into 9000 and 1000 sequences for training and testing respectively. Using Adam optimizer with a learning rate of $2 \times 10^{-3}$ and a batch size of 100, we train all the models, each composed of 10 layers, for 30 epochs. The loss function $\mathcal{L}$ is kept as
\begin{gather}\label{t19}
\mathcal{L} = \frac{1}{2} \biggl\{ f(L_i, \overline{L_i}) +  f(S_i, \overline{S_i}) \biggr\},\\
f(X_i,\overline{X_i}) = \frac{1}{N}\sum\limits_{i = 1}^N \|X_i -\overline{X_i} \|_F^2, \notag
\end{gather}where $f(X_i,\overline{X_i})$ is the mean-square error (MSE) loss function, $N$ is the total number of training sequences in the dataset, and $\left\{ {{{\overline{L}}_i},{{\overline{S}}_i}}\right\}_{i = 1}^N$ are the low-rank and the sparse components predicted by the network against their groundtruths  $\left\{ {{{L}_i},{{S}_i}}\right\}_{i = 1}^N$. Network configurations such as  kernel size and weight initialization are kept identical to \cite{8836615}, for a fair comparison.
\subsection{Results and Analysis}
\begin{figure*}
% 		\hspace{-0cm}
% 		\centering
%         \captionsetup{justification=centering}
                    \begin{subfigure}{0.325\textwidth}
%     \captionsetup{justification=centering}
   \includegraphics[width=1\linewidth]{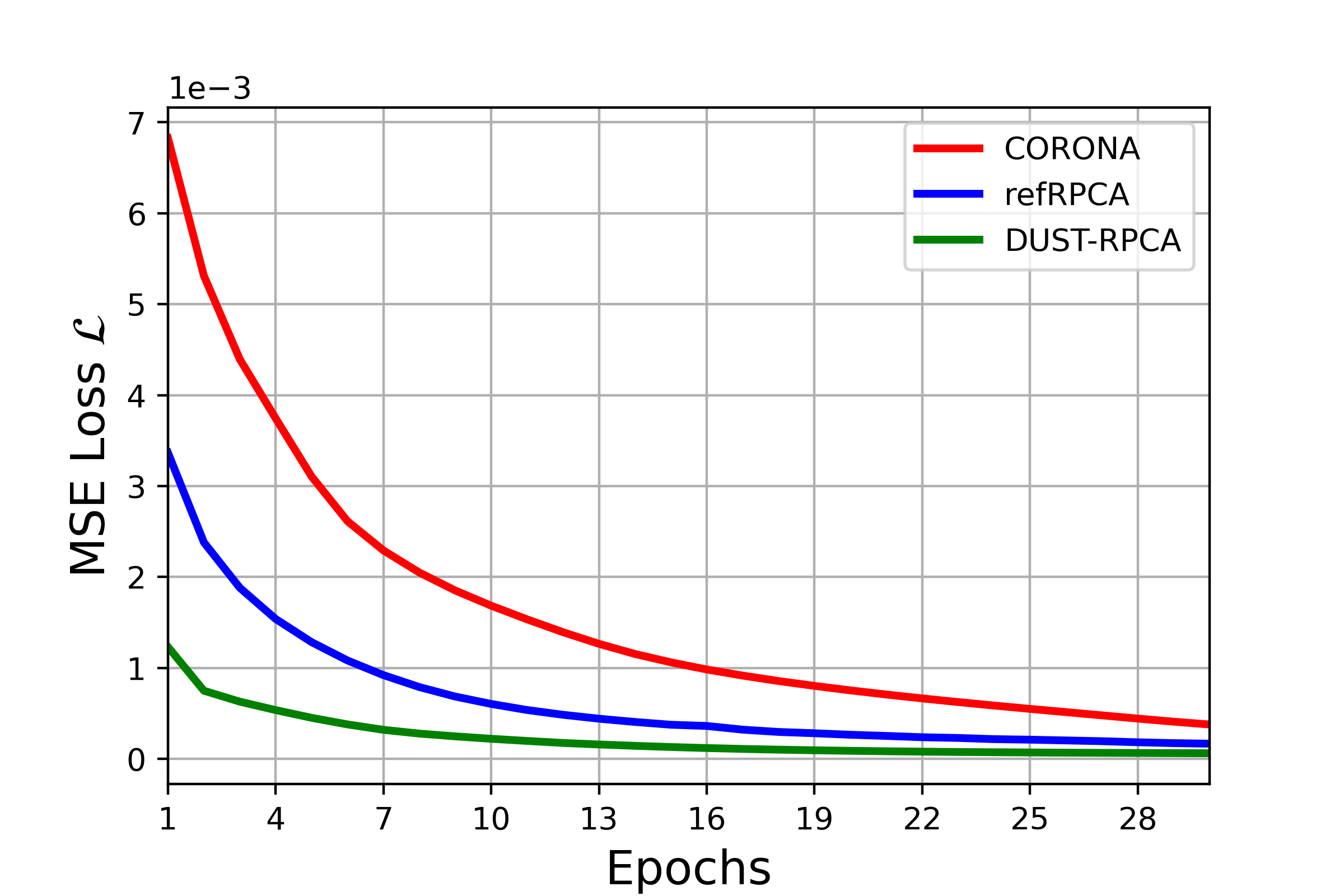}
    \caption{}
   \label{fig:2a}
  \end{subfigure}
         \begin{subfigure}{0.325\textwidth}
%     \captionsetup{justification=centering}
   \includegraphics[width=1\linewidth]{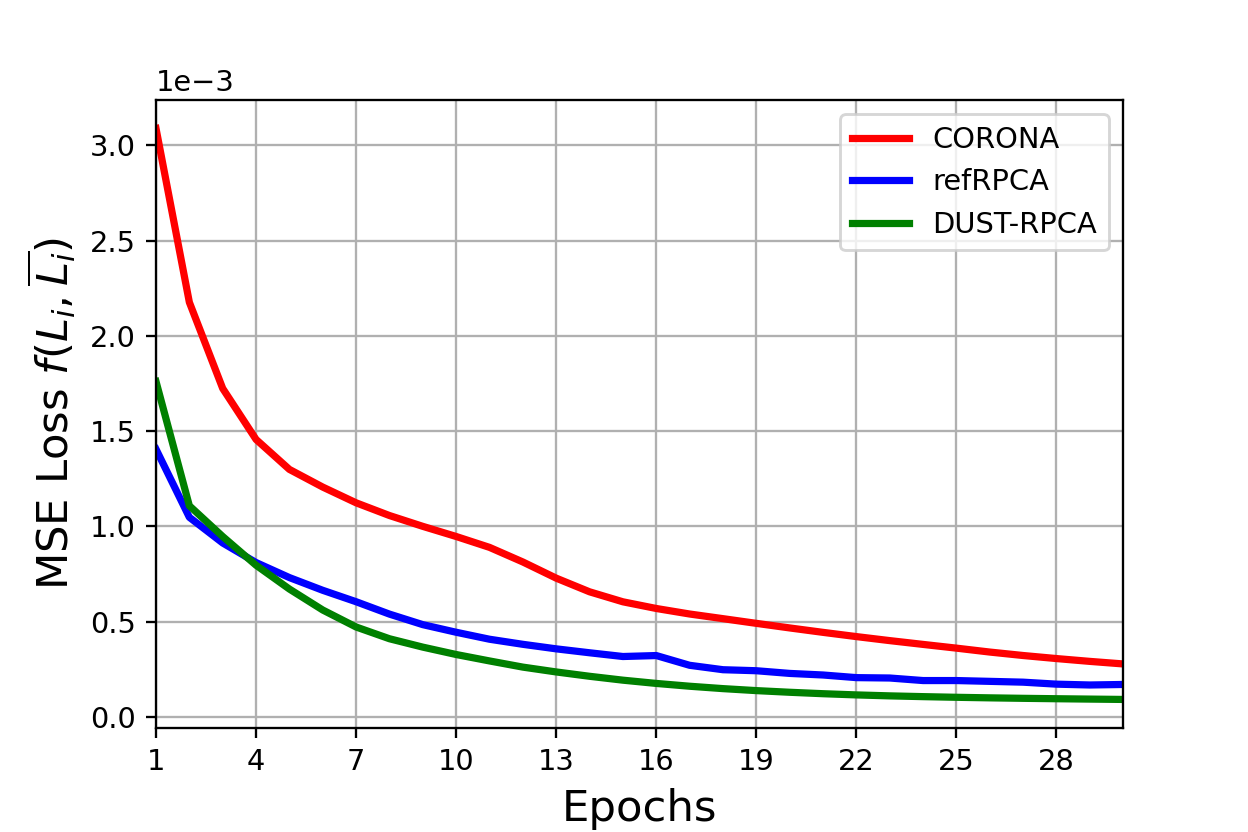}
    \caption{}
   \label{fig:2b}
  \end{subfigure}
           \begin{subfigure}{0.325\textwidth}
%     \captionsetup{justification=centering}
   \includegraphics[width=1\linewidth]{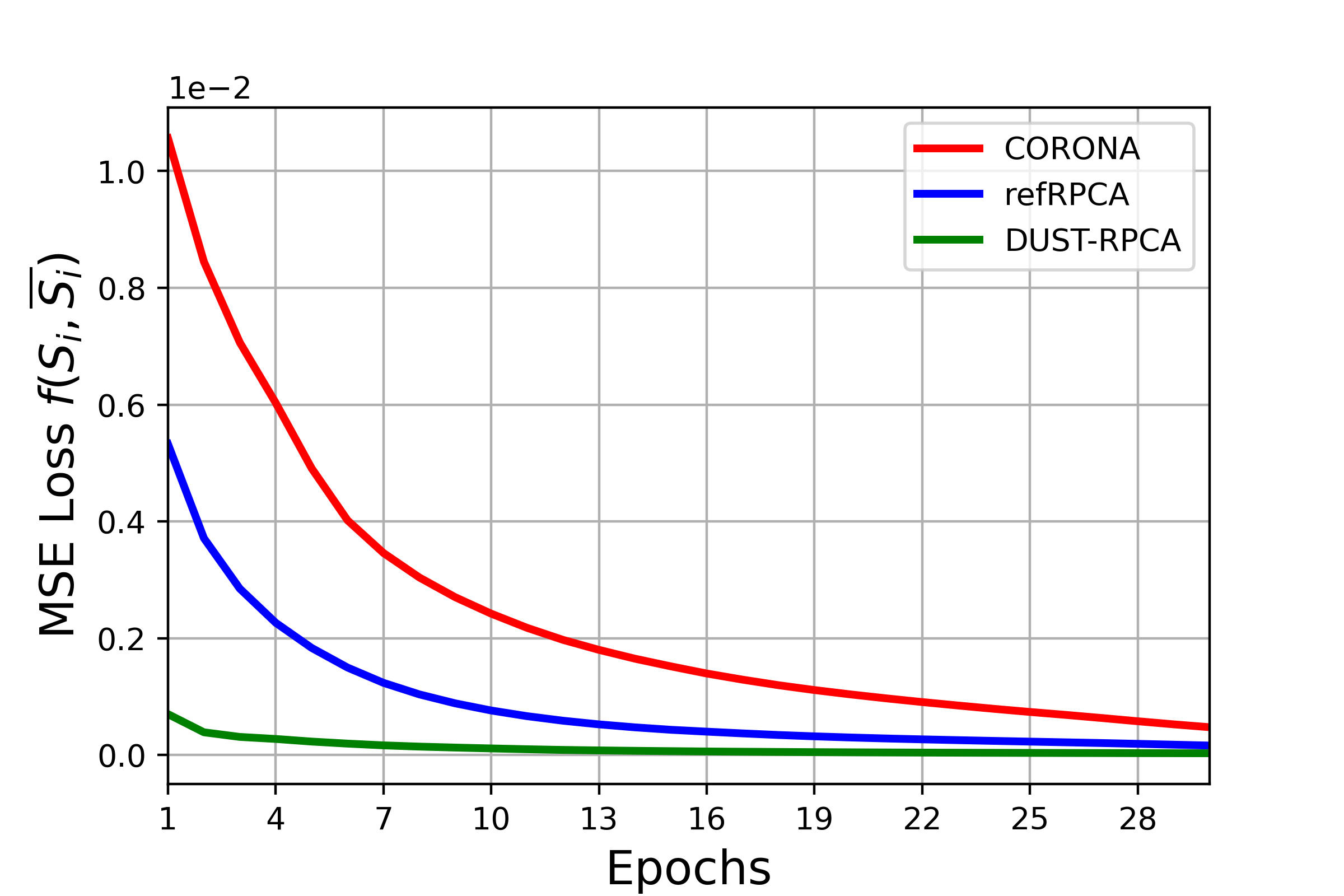}
    \caption{}
   \label{fig:2c}
  \end{subfigure}
		\caption{Comparison of deep unfolded RPCA methods of CORONA, refRPCA and DUST-RPCA: (a) MSE loss $\mathcal{L}$ (b) $ f(L_i, \overline{L_i})$ loss (c) $f(S_i, \overline{S_i})$ loss. } 
   \label{results comparision}
	\end{figure*}

Fig. \ref{fig:3} shows the visual separation results of DUST-RPCA on a sample frame from the testing dataset sequences. DUST-RPCA accurately recovers both the foreground and the background. 
%The results show that the average MSE  achieved by CORONA, refRPCA and DUST-RPCA on the testing dataset plotted against the number of epochs. 
%In addition, Fig. \ref{fig:2b} and Fig. \ref{fig:2c} show the corresponding  MSE for the sparse and low-rank components, $ f(L_i, \overline{L_i})$ and $f(S_i, \overline{S_i})$ respectively. 
The results presented in Fig. \ref{results comparision} show that the DUST-RPCA achieves significantly lower MSE for both the low-rank and sparse components, as well for the aggregate, than that achieved by other networks. It also converges much earlier than other networks. 
The refRPCA, building upon the framework of the CORONA network, formulates its problem as
\begin{gather}\label{t20}
  \min_{L,S} \quad\frac{1}{2}\|D-H_1L-H_2S \|_F^2 + \eta_1\|L\|_* \notag\\ +\eta_2\|Q\circ S\|_1 + \eta_3\|Q\circ (S-S_P)\|_1,
 \end{gather}
where $S_P$ represents the matrix containing reference frames, $Q$ denotes the weighting matrix and  $\eta_1, \eta_2$ and $\eta_3$ are the tuning regularizers. $S_P$ is constructed as $S_P = [s_1, Ps_1,  ....Ps_{q-1}]$, where $s_q$ represents the $q^{\text{th}}$ column of the corresponding sparse matrix $S$ and $P$ represents the projection matrix. However, it makes the simplifying assumption that the projection matrix $P$ does not significantly vary from one frame to the other. As such, it cannot fully incorporate the more complex correlation between frames. More importantly, both $P$ and $Q$, which are learned by the network during training, remain invariant at inference time to the data matrix $D$, which may vary significantly from one input sample to another. The proximal operator of DUST-RPCA, on the other hand, adapts for every input at each layer of the network, making it more robust to the varying input data sequences. Its performance would hold even in applications such as cloud removal, where there is little to no correlation between the successive sparse foregrounds in $S$. Although the correlation between successive sparse foregrounds may vary from one application to the other, the correlation between the background frames remains high. In fact, RPCA makes use of this phenomenon in its $D=L+S$ model to solve the background-foreground separation problem. However, amidst noise and other phenomenona such as occlusion, clutter and jitter, the high correlation between background frames starts to decrease. To solve this problem, unlike other deep unfolded networks, DUST-RPCA jointly learns the continuity across both spatial and temporal dimensions of the background thus achieving better MSE performance than both CORONA and refRPCA.

Table \ref{tab:template} shows the average MSE   achieved on the testing dataset by CORONA, refRPCA and DUST-RPCA along with the average inference time, in seconds, per frame of the MNIST sequence. While DUST-RPCA achieves the lowest MSE, it also takes more time at inference due to the one-time computation of $A_s$ and $A_t$ for every input and the addition of more convolutional layers. However, it is the SVD computation that makes up the computational bottleneck for DUST-RPCA much like CORONA and refRPCA. Moreover, the inference time is still much less than that of traditional RPCA optimization methods which use relatively more layers/iterations than the model-aided RPCA networks \cite{8836615}.
\begin{table}[t]
\centering
\begin{tabular}{|l|c|c|c|c|c|}
\hline
Method & MSE  $ (\times 10^{-4})$ & Time in seconds $(\times 10^{-3})$\\
\hline
 CORONA & 3.776 &  1.190 \\
\hline
 refRPCA & 1.668 & 2.475 \\
\hline
 DUST-RPCA & 0.614 & 5.381\\
\hline
\end{tabular}
\caption{Accuracy and efficiency comparison of DUST-RPCA with other deep unfolded RPCA networks.}
\label{tab:template}
\end{table}\section{Conclusion}
Deep unfolded networks have demonstrated improved convergence and robustness over  their iterative counterparts. Unlike other unfolded RPCA networks, the prosposed DUST-RPCA scheme incorporates spatial and temporal consistency constraints by way of regularized graph Laplacian matrices achieving better accuracy. It also reweights the thresholds for the proximal operator for every input at each layer of the network making the proposed DUST-RPCA more adaptive. Moreover, the proposed DUST-RPCA can also be used in applications, for instance medical imaging, where there may be little to no correlation between foregrounds in successive frames. 
% \section*{References}
%======================================================================== %		https://www.overleaf.com/project/623f06b4461d89875d79b888
    \balance
    \bibliographystyle{unsrt}
	\bibliography{references}
	
%======================================================================== %		
%%-------------------------------------------------------------------%%

\end{document}